\documentclass[conference]{IEEEtran}
\IEEEoverridecommandlockouts
\usepackage{cite}
\usepackage{amsmath,amssymb,amsfonts}
\usepackage{algorithmic}
\usepackage{graphicx}
\usepackage{textcomp}
\usepackage{xcolor}
\usepackage{bm}
\usepackage{multirow}
\usepackage[ruled,linesnumbered,boxed]{algorithm2e}  
\usepackage{booktabs}
\usepackage{amsmath}
\def\BibTeX{{\rm B\kern-.05em{\sc i\kern-.025em b}\kern-.08em
    T\kern-.1667em\lower.7ex\hbox{E}\kern-.125emX}}

\begin{document}
\DeclareRobustCommand*{\IEEEauthorrefmark}[1]{
    \raisebox{0pt}[0pt][0pt]{\textsuperscript{\footnotesize\ensuremath{#1}}}}
\bibliographystyle{IEEEtran}
\title{GTC: GNN-Transformer Co-contrastive Learning for Self-supervised Heterogeneous Graph Representation\\
\thanks{\IEEEauthorrefmark{*} Corresponding author.}
}
\author{Yundong Sun\IEEEauthorrefmark{1}, Dongjie Zhu\IEEEauthorrefmark{2,*}, Yansong Wang\IEEEauthorrefmark{2}, Zhaoshuo Tian\IEEEauthorrefmark{1},~\IEEEmembership{Member,~IEEE} \\

\IEEEauthorrefmark{1}Department of Electronic Science and Technology, Harbin Institute of Technology, Harbin, China \\
\IEEEauthorrefmark{2}Faculty of Computing, Harbin Institute of Technology, Weihai, China\\
Email: hitffmy@163.com, zhudongjie@hit.edu.cn, yansongwang0629@163.com,\\fuyansheng123@gmail.com, tianzhaoshuo@126.com
}
\maketitle
\begin{abstract}
Graph Neural Networks (GNNs) have emerged as the most powerful weapon for various graph tasks due to the message-passing mechanism’s great local information aggregation ability. However, over-smoothing has always hindered GNNs from going deeper and capturing multi-hop neighbors. Unlike GNNs, Transformers can model global information and multi-hop interactions via multi-head self-attention and a proper Transformer structure can show more immunity to the over-smoothing problem. So, \textit{can we propose a novel framework to combine GNN and Transformer, integrating both GNN's local information aggregation and Transformer's global information modeling ability to eliminate the over-smoothing problem?} To realize this, this paper proposes a collaborative learning scheme for GNN-Transformer and constructs GTC architecture. GTC leverages the GNN and Transformer branch to encode node information from different views respectively, and establishes contrastive learning tasks based on the encoded cross-view information to realize self-supervised heterogeneous graph representation. For the Transformer branch, we propose Metapath-aware Hop2Token and CG-Hetphormer, which can cooperate with GNN to attentively encode neighborhood information from different levels. As far as we know, this is the first attempt in the field of graph representation learning to utilize both GNN and Transformer to collaboratively capture different view information and conduct cross-view contrastive learning. The experiments on real datasets show that GTC exhibits superior performance compared with state-of-the-art methods. Codes can be available at https://github.com/PHD-lanyu/GTC. 
\end{abstract}


\section{Introduction}

The great success of Graph Neural Networks (GNNs) is attributed to the powerful local information aggregation ability of its message-passing mechanism\cite{zhang2022graph,xia2022substructure,zhang2019heterogeneous}. This mechanism fully captures the graph property that nodes and their immediate neighbors have a high probability of belonging to the same category.
  However, with the deepening research on GNNs, some inherent problems have been gradually unveiled, of which over-smoothing is the most headache problem. This leads to the fact that the existing GNNs cannot go as deep as Convolutional Neural Networks (CNNs), generally within 3-4 layers\cite{li2019deepgcns,bose2023can}, so there is a bottleneck when capturing multi-hop neighbors. There is a growing recognition that multi-hop neighborhood information is the key to capturing global structural information, and failure to effectively capture multi-hop neighbors will limit the model’s learning capacity\cite{9807384}. Moreover, some studies\cite{zhu2020beyond} have shown that in heterophilic graphs, the labels between immediate neighbors are often different, thus requiring multi-hop interactions between nodes. Over-smoothing is not only a key limitation in developing deep GNNs, but also seriously deteriorates the ability of GNNs to handle heterophilic graphs. Therefore, how to solve the limitations of the message-passing mechanism in GNNs, eliminating (or at least alleviating) the over-smoothing problem, and thus improving the expressive ability of the model is an urgent problem in this field.
\begin{figure}[t]
  \centering
  \includegraphics[width=3.5in,trim=0 20 0 20]{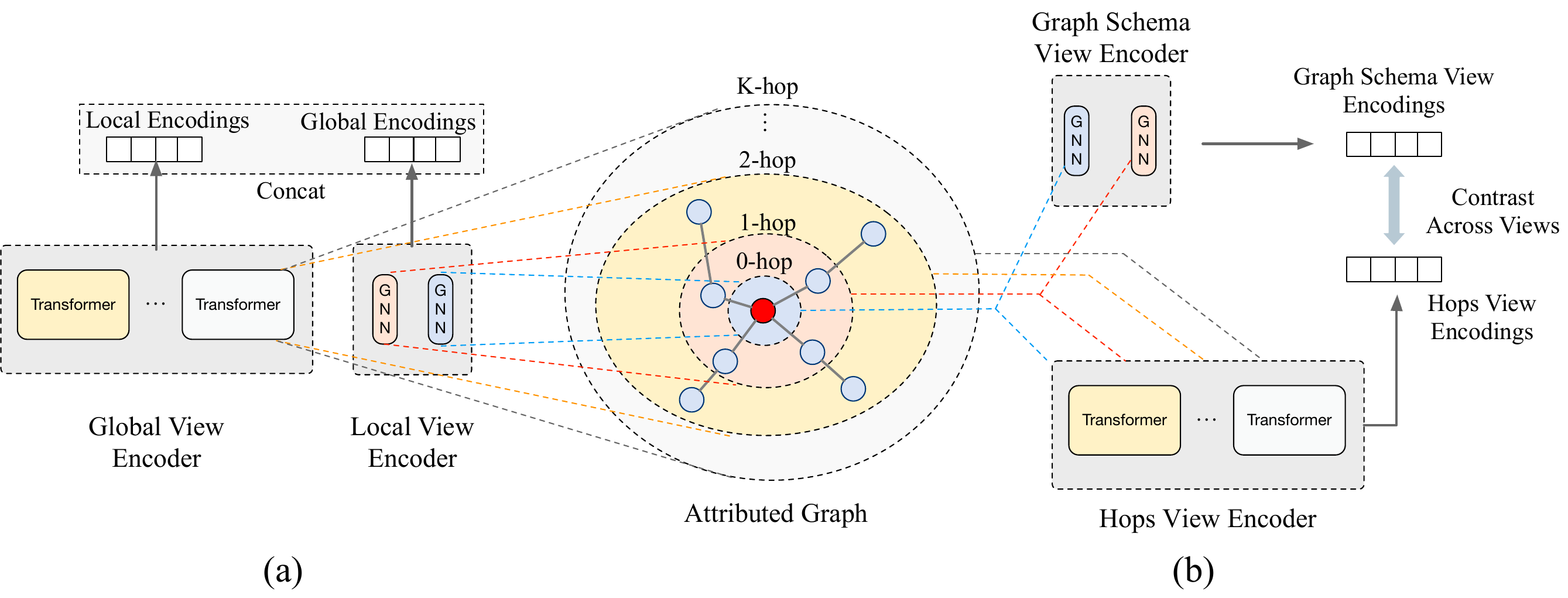}
  \caption{The diagram of (a) local view and global view, and (b) graph schema view and hops view collaborate encoding by GNN and Transformer.}
  \label{fig1}
  \vspace{-0.5cm}
  \end{figure}
  
Based on the Multi-head Self-Attention(MSA) mechanism, Transformer models have already shown dominant performance in the fields of natural language processing(NLP)\cite{wu2023graph,9835527}, computer vision(CV)\cite{DosovitskiyB0WZ21,pan2022hilo}, and vision-language pretraining (VLP)\cite{li2021align,li2022blip}. The advantage of Transformer is that it can deeply mine long-range full interactions. If it is applied to graph representation learning, it can well model the global graph structure information. At the same time, some researchers have shown that even deep Transformer structures can show immunity to over-smoothing\cite{chen2023nagphormer,li2022adaptive}.

Some research in CV leverages CNNs or Multilayer Perceptron(MLP) to extract local information and capture global information by Transformer, which has achieved significant performance improvement\cite{pan2022less,xiao2021early}. Inspired by this idea, we raise the first question of this paper: \textbf{\textit{Can we leverage GNN and Transformer as two branches to collaboratively encode the local view and global view (as shown in Fig.\ref{fig1}(a)) or graph schema view and hops view (as shown in Fig.\ref{fig1}(b)), respectively, to exploit the local information aggregation capability of GNN and Transformer’s global structure modeling capability to avoid the over-smoothing problem?}} This is the first challenge that needs to be overcome in this paper.

On the other hand, although GNNs have achieved promising performance on several graph tasks, most of the existing methods\cite{wang2019heterogeneous,hamilton2017inductive,zeng2022accurate} belong to the semi-supervised learning paradigm. In this paradigm, some labeled nodes are indispensable during the training stage\cite{zhang2022unsupervised}.
However, obtaining node labels in some real application scenarios is extremely challenging and time-consuming. For example, accurately labeling unknown protein molecules requires a lot of professional domain knowledge, which is also challenging for professionals\cite{li2022graph}. In addition, semi-supervised learning also encounters poor generalization and weak robustness problems\cite{liu2022graph}. Therefore, realizing self-supervised heterogeneous graph representation learning is of great significance to improve the model's applicability in real scenarios.  

In the past two years, the research on graph contrastive learning(GCL) has become the most concerned topic in graph representation learning. GCL first encodes node or graph representations under different views through different branches or channels. Then, based on the different view representations, different contrastive tasks are constructed\cite{wang2021self,liu2023hierarchical}. The goal is to maximize the similarity of representations obtained from the same graph or node across different views while minimizing the similarity of representations between different graphs or nodes. Due to the particularity of the graph structure, it is easy to obtain different view information through diverse strategies (such as random neighbor sampling). Therefore, GCL has become a preferred solution for self-supervised graph learning.
  
Based on the above analysis, we raise the second question of this paper: \textbf{\textit{Can we construct a collaboratively contrastive learning task based on the different view representations obtained by GNN and Transformer, boosting the information fusion of the two views, and realizing self-supervised heterogeneous graph representation learning?}} This is the second challenge that needs to be overcome in this paper.
  
  With the above two questions, we propose a GNN-Transformer collaborative learning scheme. Through collaborative learning between GNN and Transformer, we can achieve efficient, anti-over-smoothing, and self-supervised heterogeneous graph learning. The contributions of this paper can be summarized as follows:

\begin{itemize}
\item We propose a novel scheme that leverages GNN and Transformer as two branches to encode different view information and perform cross-view collaboratively contrastive learning. This scheme breaks through GNNs' limitation of capturing multi-hop neighbor information on the premise of avoiding the over-smoothing problem. As far as we know, this is the first attempt in this field. 
\item We construct the GNN-Transformer Co-contrastive learning architecture, abbreviated as GTC. In GTC, the GNN and Transformer are leveraged as two branches to encode graph schema view and hops view information respectively, and the contrastive learning task is established based on the encoded cross-view information to realize self-supervised heterogeneous graph representation.
\item For the Transformer branch, we first propose Metapath-aware Hop2Token, which realizes the efficient transformation from different hop neighbors to token series in heterogeneous graphs. The proposed CG-Hetphormer model can not only attentively fuse two-level semantic information, Token-level and Semantic-level, but also cooperates with the GNN branch to achieve collaborative learning.
\item We have conducted tremendous experiments on real heterogeneous graph datasets. The experimental results show that the performance of our proposed method is superior to the existing state-of-the-art methods.
In particular, when the model goes deeper, our method can also maintain high performance and stability, this proves that the multi-hop neighbor information can be captured without interference from the over-smoothing problem, which provides a reference for future research on solving the over-smoothing problem of GNNs.
\end{itemize}

\section{Related Work}\label{sec2}

\subsection{Works to Resist GNNs' Over-smoothing}\label{sec2.1}


One type of method is dedicated to alleviating over-smoothing by adding or modifying some substructures (such as normalization layer, residual structure.)\cite{li2019deepgcns,li2020deepergcn,guo2023contranorm}. 
Another type of method is devoted to incorporating some over-smoothing measures (e.g., Dirichlet energy, Mean Average Distance.) to constrain or guide the training process of GNNs for the purpose of mitigating over-smoothing\cite{zhou2021dirichlet,chen2020measuring}. 

The above two types of methods have made great progress in mitigating GNNs' over-smoothing and significant contributions to the deep GNN models. However, all approaches require modifications to the GNN's model structure or introducing additional training constraints. Differing from the above two types of methods, this paper is not limited to optimizing the GNN itself but instead leverages Transformer's ability to capture multi-hop interactions to make up for GNN's shortcomings and allow GNNs to focus on local information extraction. This paper is a more novel approach and orthogonal to existing methods, and can be combined with them for better performance.

\subsection{Graph Contrastive Learning}\label{sec2.2}


Graph augmentation and contrastive tasks are the two core modules of GCL. Among the existing methods, the graph augmentation methods mainly include three categories: attribute-based\cite{jin2021multi,hu2020strategies}, topological-based\cite{wang2021self,liu2023hierarchical} 
and hybrid augmentations\cite{jiao2020sub,zhu2021graph}. 
Different contrastive tasks can enrich supervision signals from instances with similar semantic information. Following the review\cite{liu2022graph}, the existing works can be divided into two categories: same-scale and cross-scale methods. Different branches of the same-scale approaches\cite{verma2021towards,you2020graph} discriminate instances at the same scale (e.g., node vs node), while the cross-scale approaches\cite{velickovic2019deep,park2020unsupervised} conduct comparison at different granularities (e.g., node vs graph). 


The methods mentioned above have promoted the progress of GCL, making graph representation learning continue to develop in the direction of self-supervised learning so that it can be more widely used in various graph applications. 
Although this paper leverages different branches to encode information from different views and then establishes the GCL task, which follows the paradigm of the common GCL. However, this paper innovatively regards the graph schema view and hops view as different branches, maximizing the advantages of GNN and Transformer. More importantly, the information of multi-hop neighbors can be captured while avoiding over-smoothing, which is the most remarkable superiority to the existing methods.

\subsection{Transformer for Graph}\label{sec2.3}


Graph data usually contain more complex attributes, including both node features and structural topology, which cannot be directly encoded into Transformer's tokens\cite{chen2023nagphormer}. This leads to Transformer cannot directly model the topology of the graph. Existing methods address this issue by introducing a structural encoding module\cite{kreuzer2021rethinking}, leveraging GNN as an auxiliary module\cite{wu2021representing}, or incorporating graph bias into the attention matrix\cite{ying2021transformers}, making Transformer more competitive in graph analysis tasks.
However, most of the existing methods take the entire graph as input and treat each node as a token for subsequent calculations. The quadratic complexity of the MSA makes the model unable to handle large-scale graphs.
 Therefore, some of the latest research is devoted to optimizing the computational efficiency of Transformer so that it can perform analysis tasks on large-scale graphs\cite{chen2023nagphormer,wu2022nodeformer}.

Although some methods\cite{kreuzer2021rethinking,wu2021representing} solve the shortcomings of Transformer in graph structure encoding to some extent, NAGphormer\cite{chen2023nagphormer} also realizes mini-batch training on large-scale graphs. However, most of the existing methods stay in homogeneous graphs and belong to the category of supervised learning. Our well-designed Metapath-aware Hop2Token is capable of transforming different hop neighbors into token series in heterogeneous graphs. The proposed CG-Hetphormer model can attentively fuse two-level semantic information. Meanwhile, our method is oriented to self-supervised GCL without any node labels. These are the two notable differences between this paper and the existing methods.

\section{Preliminaries}\label{sec3}


\subsection{Notations}

\textbf{Heterogeneous Graph.} 
The difference between heterogeneous and homogeneous graphs is that there are multiple types of nodes and edges in heterogeneous graphs while homogeneous graphs treat all nodes and edges as one type.
 If the heterogeneous graph is denoted as $G=\left(V,E\right)$, then $type(V)|+|type(E)|>2$, where $V$ and $E$ represent node and edge set, respectively. For a general homogeneous graph, we denote the feature matrix of the node as $X\in\mathbb{R}^{n\times d}$, and the vector $x_i\in\mathbb{R}^{1\times d}$ formed by the $i$-th row of $X$ is the feature of the node $i$. As the adjacency matrix of $G$, $A\in\mathbb{R}^{n\times n}$ represents the affinity or immediate interaction between nodes in the graph. For example, $A_{ij}=1$ denotes an edge between nodes $i$ and $j$, that is, there is an immediate interaction between them. For heterogeneous graphs, due to various types of nodes and edges, there may be multiple node feature and adjacency matrices. 
Compared with homogeneous graphs, heterogeneous graphs can meet more requirements of real application scenarios and exhibit more substantial semantic capabilities. But it also brings more significant challenges to semantic representation and analysis.

\textbf{Metapath.} Metapath is a path with a special semantic relationship in a heterogeneous graph. It is usually expressed as ${v_1}\mathop  \to \limits^{{R_1}} {v_2}\mathop  \to \limits^{{R_2}} ...\mathop  \to \limits^{{R_l}} {v_{l + 1}}$, abbreviated as $v_1v_{2\ldots}v_{l+1}$, which represents nodes $v_1$ and $v_{l+1}$ has a composite relation $R=R_1\circ R_2\circ\ldots R_l$. Compared with the immediate interaction in the adjacency matrix $A$, the metapath can represent indirect interactions with multi-hop neighbors that contain richer semantic information.

\textbf{Metapath-based Multi-Hop Neighbors.} In a heterogeneous network, given a node $i$ and metapath $\varphi$, starting from node $i$, we denote the set of nodes $\mathcal{N}_\varphi^k\left(i\right)$ reached along metapath $\varphi$ through $k$ hops as the $k$-hop neighbors of node $i$ under metapath $\varphi$.

\subsection{Problem Definition}

For the general node classification task, we need some nodes $V_l$ with labels $Y_{V_l}$ as the training set. We need to leverage the feature $X_{V_l}$ of the training set node as input and $Y_{V_l}$ as the label to train the model. After the training is completed, the model is used to predict the label of the remaining unlabeled nodes $V_u$. For the self-supervised representation learning in this paper, there is no need for any labels during the training phase. We can obtain a deployable model simply by using the feature $X_{V_l}$ of the training nodes. Then the deployed model can be leveraged to get the node embedding which is used for downstream tasks. Node clustering task is similar to classification task, but it is unsupervised generally. In this paradigm, we first use the deployed node representation model to obtain node embedding and then input the embedding into node clustering algorithms, like K-means, to finish the task.

\subsection{Graph Neural Networks}
\begin{figure*}[t]
  \centering
  \includegraphics[width=6.5in,trim=0 30 0 40]{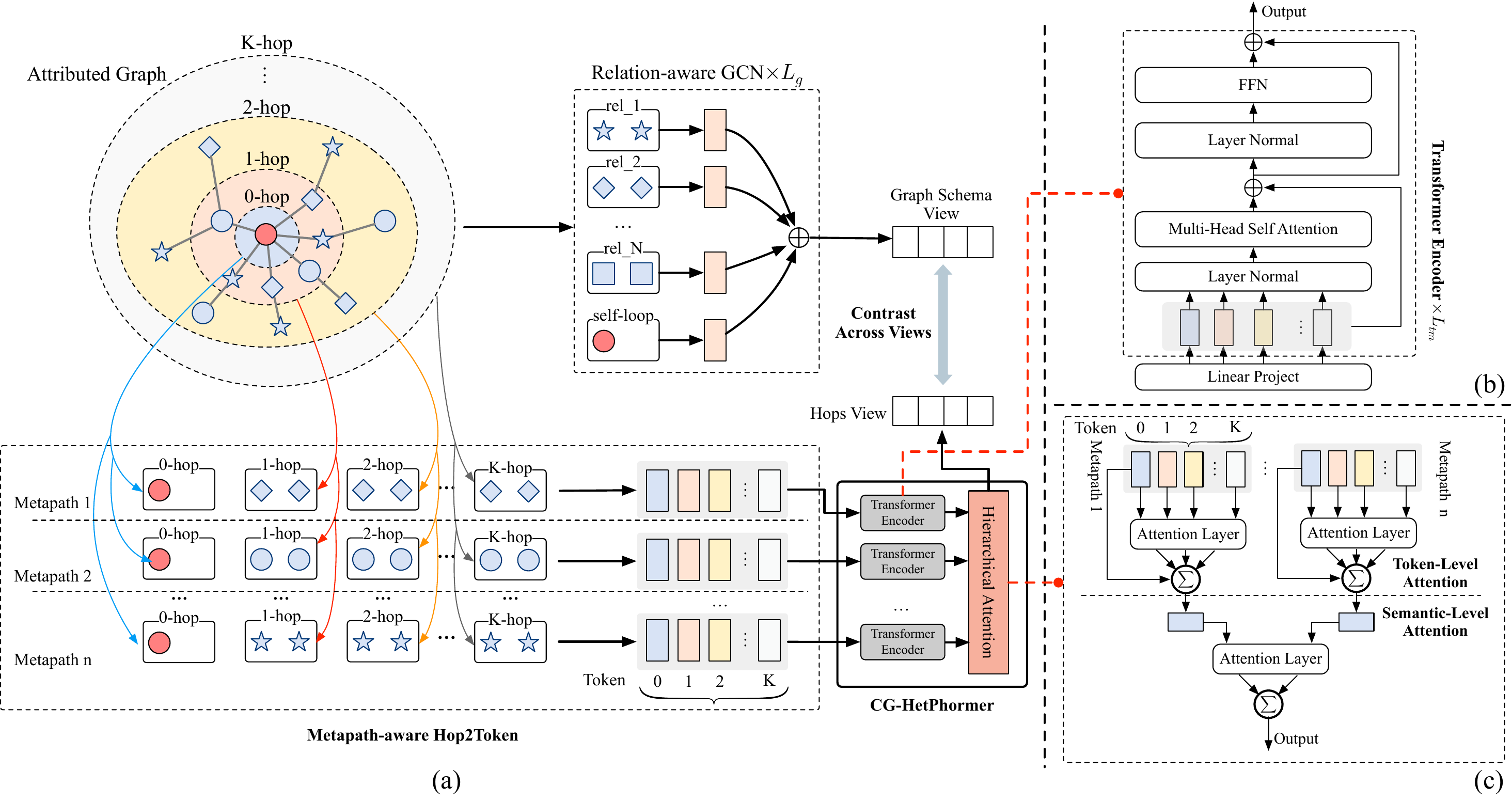}
  \caption{(a) is the model framework of GTC, (b) and (c) are the detailed structure of the Transformer Encoder and Hierarchical Attention respectively.}
  \label{fig2}
  \vspace{-0.5cm}
  \end{figure*}
The essence of GNNs is to utilize message-passing operations at each layer to aggregate information from immediate neighbors. 
 Here, the operations in each layer of the network can be expressed as follows:
\small{
\begin{equation}
	H^{\left(l+1\right)}=\sigma\left(\left(D+I\right)^{-1/2}\left(A+I\right)\left(D+I\right)^{-1/2}H^{\left(l\right)}W^{\left(l\right)}\right)
\end{equation}
}
where $H^{\left(l\right)}\in\mathbb{R}^{n\times d^{\left(l\right)}}$, $d^{\left(l\right)}$ and $W^{\left(l\right)}\in\mathbb{R}^{d^{\left(l\right)}\times d^{\left(l+1\right)}}$ denotes the node representations, hidden dimension and learnable parameter matrix of the $l$-th layer network, respectively. $n$ is the number of nodes, $D$ is the degree matrix of adjacency matrix $A$, $I\in\mathbb{R}^{n\times n}$ is the identity matrix, which is utilized to add self-loops for each node at each layer. $\sigma$ is the nonlinear activation function.

\subsection{Transformer}

In the original design, a complete Transformer architecture contains two parts, the encoding and decoding structure, but there is only the encoding structure of the Transformer in this paper. In general, an encoding structure contains multiple Transformer layers, in which the MSA substructure is the core module. MSA is utilized to capture the semantic correlation between input tokens and is the cornerstone of the Transformer's powerful representation capability. Since MSA can be regarded as a concatenation of various single self-attention headers, for simplicity, we describe it with one header. The self-attention module will first map the input token feature $H\in\mathbb{R}^{n\times d}$ into three subspaces, namely $Q$, $K$, and $V$:

\begin{equation}
Q=HW^Q,K=HW^K,V=HW^V	
\end{equation}
where $W^Q\in\mathbb{R}^{d\times d_K}$, $W^K\in\mathbb{R}^{d\times d_K}$ and $W^V\in\mathbb{R}^{d\times d_V}$ are learnable mapping matrices, $d$ is the dimension of token feature, $d_K$ is the hidden dimension of $Q$ and $K$, $d_V$ is the hidden dimension of $V$, $n$ is the number of tokens. After leveraging the three subspace features to calculate pairwise interactions between the input tokens, row-wise softmax is applied to obtain the output representation:
\begin{equation}
	H^\prime=softmax\left(\frac{QK^\top}{\sqrt{d_K}}\right)V
\end{equation}

\section{The Proposed Methodology}\label{sec4}

In this section, we will introduce our proposed method in detail. To eliminate the over-smoothing problem and realize the self-supervised heterogeneous graph representation by collaboratively learning of GNN and Transformer. This paper proposes the GTC architecture, which is the abbreviation of GNN-Transformer Co-contrastive learning for self-supervised heterogeneous graph representation. As shown in Fig.\ref{fig2}, under the GTC architecture, a contrastive learning task is established based on the embedding from the cross-view of the graph schema view and hops view.
 For the Transformer branch, we propose the Collaborate-GNN Heterogeneous Transformer model (CG-HetPhormer), which can leverage the Transformer Encoder and the proposed Hierarchical Attention structure to cooperate with GNN, realizing more efficient encoding of neighborhood information from different hops and different metapaths.

\subsection{Node Feature Transformation}\label{sec4.1}

In a heterogeneous graph, the feature dimensions of different types of nodes are inconsistent, and these features are located in disparate spaces. Before we utilize the node features, we need to map the features from these different spaces into a unified space and realize the unification of dimensions. One intuitive way is to create a mapping matrix for each node type to accomplish this task:
\begin{equation}
	h_{i}=\sigma \left( W_{\varphi_{i} }\cdot x^{\top }_{i}+b_{\varphi_{i} }\right)
\end{equation}
where $x_i\in\mathbb{R}^{1\times d_{\varphi_i}}$ is the raw feature of node $i$, $\varphi_i$ represents the type of node $i$, $d_{\varphi_i}$ is the raw feature dimension of $\varphi_i$ type nodes, $W_{\varphi_i}\in\mathbb{R}^{d\times d_{\varphi_i}}$ is the mapping matrix for $\varphi_i$ type nodes, $b_{\varphi_i}$ is the corresponding bias vector, $d$ is the dimension of the target space, $\sigma$ is the activation function.

\subsection{Graph Schema View Branch}\label{sec4.2}
In this section, our objective is to obtain the representation of nodes under the graph schema view, as shown in the top of Fig.\ref{fig2}(a). To better deal with different types of nodes and different relationships in heterogeneous graphs, we adopt the simple yet effective information aggregation scheme similar to RGCN\cite{schlichtkrull2018modeling}:

\begin{equation}
\begin{split}	
	h_i^{\left(l+1\right)}=\sigma & \left(\sum_{r\in\mathcal{R}}\sum_{j\in N_i^r}{\frac{1}{c_{i,r}}}W_r^{\left(l\right)}h_j^{\left(l\right)}+W_0^{\left(l\right)}h_i^{\left(l\right)}\right) \\ & h_i^{\left(0\right)}=h_i,Z_i^{sv}=h_i^{\left(L_g\right)}
	\end{split}
\end{equation}


where $L_g$ is the number of network layers, $N_i^r$ denotes the immediate neighbors of node $i$ under the relationship $r$. $c_{i,r}$ is a normalization constant, which can be $c_{i,r}=\left|N_i^r\right|$, or a learnable parameter. $h_i$ is the projected node feature derived from section \ref{sec4.1}.

After the $L_g$ layers, we can obtain the representation $Z_i^{sv}$ of node $i$ under the graph schema view. By this scheme, the information aggregation of immediate neighbors can be well realized under different node relationships, and the self-loop is added so that it can well maintain the original features. In addition, thanks to the simple structure, the computational efficiency is satisfactory.

It should be noted that the purpose of this paper is to establish collaboratively contrastive learning between graph schema view and hops view, and finally, it can solve GNNs' over-smoothing problem while capturing multi-hop neighbors. The model structure and information aggregation scheme of the graph schema view branch is not our focus, so it can be replaced by any other methodologies of GNNs.

\subsection{Hops View Branch}\label{sec4.3}
\subsubsection{Metapath-aware Hop2Token}\label{sec4.3.1}
 
One of the keys to applying Transformer to graph data is how to transform graph information (including node feature and graph structure information) into tokens. Some existing methods treat each node as a token and resort to different methods (such as Laplacian eigenvectors) to introduce graph structure information\cite{kreuzer2021rethinking}. This approach would lead to high computational complexity, which limits the model to small-scale graph data. Inspired by the Hop2Token of NAGphormer\cite{chen2023nagphormer}, this paper also adopts the strategy of transforming the information of the same hop neighbors into one token. The difference is that NAGphormer is originally designed for homogeneous graphs. To deal with heterogeneous graphs, this paper designs Metapath-aware Hop2Token, as shown in the left bottom of Fig.\ref{fig2}(a).

Metapath-aware Hop2Token first acquires neighbors from different hops within different metapaths. For node $i$, we denote $\mathcal{N}_\varphi^k\left(i\right)=\left\{j\in V\ \middle| d\left(i,j\right)=k\land\left(i,j\right)\in\varphi\right\}$ as $k$-hop neighbors of $i$ under metapath $\varphi$, where $d\left(i,j\right)$ represents the shortest path distance between node $i$ and $j$. Please note that the 0-hop neighbor is the node $i$ itself, i.e. $\mathcal{N}^0\left(i\right)=\left\{i\right\}$. After obtaining the different hop neighbors of the nodes, the neighbors from the same hop are regarded as a group, and the information aggregation operation is performed within every group.  There are various strategies (Graph Conv, SUM, MEAN, MAX, etc.)  can be selected as we need,  in this paper, our aggregation strategy is as follows:
\begin{equation}
	x_\varphi^k=({\hat{A}}_{\varphi})^{k}H,  (k=0,1,2...,K)
\end{equation}
where ${\hat{A}}_{\varphi}=D^{-1/2}A_{\varphi} D^{-1/2}$, $A_{\varphi}$ is the adjacent matrix under metapath $\varphi$. $x_\varphi^k\in\mathbb{R}^{n\times d}$ is the token representation of $k$-hop neighbors under metapath $\varphi$. $H$ is the projected feature matrix obtained from section \ref{sec4.1}, $D$ is the degree matrix of $A_{\varphi}$, $n$ is the node number and $d$ is the token dimension. Suppose we set the maximum hop to $K$, then for each node $i$, we can get a token sequence $S_{\varphi,i}=\left(x_{\varphi,i}^0,x_{\varphi,i}^1,\ldots,x_{\varphi,i}^K\right)$ with $K+1$ length.

For the other metapaths, there are consistent implementations, and we will not conduct redundant descriptions here. Finally, we can get the token sequence under each metapath. These sequences can not only represent the neighborhood information from different hops under different semantic metapaths, but also can well transform the node feature and graph structure into tokens that can be trained in a mini-batch manner, which greatly reduces the subsequent computational complexity of MSA. Most importantly, the subsequent MSA can make full use of these token sequences to capture the semantic relationship between different hop neighbors and even different metapaths. This is beyond the capabilities of existing methods.

\subsubsection{CG-HetPhormer}\label{sec4.3.2}

The overall framework of CG-HetPhormer can be seen at the right bottom of Fig.\ref{fig2}(a). After obtaining the tokens under different metapaths using the Metapath-aware Hop2Token described in \ref{sec4.3.1}, in this section, we first input the tokens under each metapath into the corresponding Transformer Encoder to further mine the semantic interactions between different hop neighborhoods under the same metapath. The proposed Hierarchical Attention module is then leveraged to conduct attentive fusion for both Token-level and Semantic-level information. Finally, we can get the representation of the nodes under the hops view. Next, we detail the Transformer Encoder and Hierarchical Attention modules, respectively.

\textbf{Transformer Encoder.} Fig.\ref{fig2}(b) depicts the detailed structure of the Transformer Encoder. It follows the encoding part of the original Transformer and removes the decoder part. We first carry out feature mapping operation for $S_{\varphi,i}=\left(x_{\varphi,i}^0,x_{\varphi,i}^1,\ldots,x_{\varphi,i}^K\right)$ via a Linear:
\begin{equation}
	Z_{\varphi,i}^{\left(0\right)}=\left[x_{\varphi,i}^0M_\varphi;x_{\varphi,i}^1M_\varphi;\ldots;x_{\varphi,i}^KM_\varphi\right]
\end{equation}
where $ M_\varphi\in\mathbb{R}^{d\times d_m}$ is the learnable mapping matrix and $Z_{\varphi,i}^{\left(0\right)}\in\mathbb{R}^{\left(K+1\right)\times d_m}$ is the projected token representation with dimension $d_m$.

In the next step, we input $Z_{\varphi, i}^{\left(0\right)}$ into $L_{tm}$ sequentially connected Transformer Encoder blocks to mine the semantic relationship between different hop neighbors. The structure of each block is consistent, including an MSA and FFN structure. Both structures contain a residual substructure, and a Layer Normal structure is inserted before the two structures. The data flow of the Transformer Encoder block is:

\begin{equation}
\begin{split}	
	&Z_{\varphi,i}^{\prime^{\left(l\right)}}=MSA\left(LN\left(Z_{\varphi,i}^{\left(l-1\right)}\right)\right)+Z_{\varphi,i}^{\left(l-1\right)} \\	 	&Z_{\varphi,i}^{\left(l \right)}=FFN\left(LN\left(Z_{\varphi,i}^{\prime\left(l\right)}\right)\right)+Z_{\varphi,i}^{\prime^{\left(l\right)}}
	\end{split}
\end{equation}
where $l=1,2\ldots,L_{tm}$ represents the index of the block. After $L_{tm}$ blocks, we can make full use of the MSA structure to realize the information interaction between different hop neighbors under the same metapath, and obtain a token sequence $Z_{\varphi,i}\in\mathbb{R}^{\left(K+1\right)\times d_m}$ representation with richer semantics. Next, to obtain the final representation of nodes, we need to perform semantic information aggregation on these tokens.

\textbf{Hierarchical Attention.} After the aforementioned procedure, we have obtained the token sequences $Z_{\varphi,i}$ of node $i$ under each metapath. We propose the Hierarchical Attention information aggregation model to better mine semantic information at different levels and obtain more representative node embeddings. It mainly includes two levels of attentive information aggregation, namely Token-level and Semantic-level.

First, at the Token-level, to better explore the importance of different tokens to the final embedding within one metapath, following NAGFormer\cite{chen2023nagphormer}, we calculate the correlation between $[1,2,...,\ K]$ hop tokens and the node itself (that is, the 0-hop neighborhood):
\begin{equation}
	\alpha_{\varphi,i}^k=\frac{exp\left(\left(Z_{\varphi,i}^0\parallel Z_{\varphi,i}^k\right)W_{\varphi,ta}^\top\right)}{\sum_{k^\prime=1}^{K}exp\left(\left(Z_{\varphi,i}^0\parallel Z_{\varphi,i}^{k^\prime} \right)W_{\varphi,ta}^\top\right)}
\end{equation}
where $W_{\varphi,ta}\in\mathbb{R}^{1\times2d_m}$ is the learnable parameter matrix, $Z_{\varphi,i}^k$ represents the $k$-th hop token of node $i$ under metapath $\varphi$. Based on this, information aggregation between different hops can be achieved:
\begin{equation}
	Z_{\varphi,i}=Z_{\varphi,v}^0+\sum_{k=1}^{K}{\alpha_{\varphi,i}^kZ_{\varphi,i}^k}
\end{equation}

After obtaining the node representations under each metapath, at the Semantic level, we also need to perform information aggregation on these node representations from different metapaths. Similarly, different metapaths express different semantics and contribute differently to the final representation of nodes on different tasks or different datasets. Therefore, we hope that the model can flexibly adjust the weight of different semantic information. First of all, we need to learn the importance of different metapaths. At the same time, to make the information fusion of each metapath meet the normalization, we use the softmax function to normalize the importance:

\begin{equation}
	\alpha_{\varphi ,i} =\frac{exp\left( \sigma \left[ \delta_{\varphi } \tanh {\left( W_{\varphi }Z_{\varphi ,i}\right)  }  \right]  \right)  }{\sum_{{\varphi }^{\prime }  \in \Phi } {exp(\sigma \left[ \delta_{{\varphi }^{\prime }  } \tanh {\left( W_{\varphi^{\prime } }Z_{{\varphi }^{\prime }  ,i}\right)  }  \right]  }  )} 
\end{equation}
where $\Phi$ is the set of metapaths, $\delta_\varphi$ and $W_\varphi$ are both learnable parameter matrixs corresponding to metapath $\varphi$ and tanh is the activation function. Finally, the node representations under the hops view is:

\begin{equation}
	Z_{i}^{hv}=\sum_{\varphi\in\Phi}{\alpha_{\varphi,i} Z_{\varphi,i}}
\end{equation}

\subsection{Collaboratively Contrastive Optimization}\label{sec4.4}

After the above process, the representations $Z_i^{sv}$ and $Z_i^{hv}$ of the node $i$ under the graph schema view and hops view are respectively obtained in section \ref{sec4.2} and \ref{sec4.3}. Before performing contrastive optimization, a key issue is how to determine positive and negative samples. One of the most intuitive strategies is only to regard the embedding of the same node under different views as positive samples, and all others as negative samples. This approach works well in fields like CV, in which the samples are independent. However, in graph data, nodes are correlated to each other, and we also try to treat highly-correlated nodes as positive samples.

In a heterogeneous graph, different metapaths represent different semantic correlations. Therefore, in this paper, we make an assumption that if there are multiple metapath instances between two nodes, it represents the highly-correlation between them. Based on this point, we first count the metapath instances between two nodes:

\begin{equation}
	C_i\left(j\right)=\sum_{\varphi\in\Phi} A_{\varphi}(i,j)
\end{equation}
We can filter out the positive samples by setting the threshold $\theta_{pos}$, that is, if $C_i\left(j\right)\geq\theta_{pos}$ we will add the node pair $(i,j)$  to the positive sample set of node $i$, if $A_{\varphi}(i,j)=1$, there is an edge between node $i$ and $j$ under the metapath $\varphi$, otherwise $A_{\varphi}(i,j)=0$.

After the above process, we expand the positive sample set, and finally get the positive sample set $\mathbb{P}$ and negative sample set $\mathbb{N}$. We can build the following function to compute the contrastive loss from the graph schema view to the hops view:

\begin{equation}
	\mathcal{L}_i^{sv}=-log{\mathrm{\Sigma}_{j\in\mathbb{P}_i}}\frac{exp{\left(sim\left(Z_i^{sv},Z_j^{hv}\right)/\tau\right)}}{\mathrm{\Sigma}_{k\in\{\mathbb{P}_i\bigcup{\mathbb{N}_i\}}}exp{\left(sim\left(Z_i^{sv},Z_k^{hv}\right)/\tau\right)}}
\end{equation}
where $sim\left(i,j\right)$ represents the cosine similarity between vector $i$ and $j$, $\tau$ is the temperature parameter. Similarly, the contrastive loss from hops view to graph schema view is:
\begin{equation}
	\mathcal{L}_i^{hv}=-log{\mathrm{\Sigma}_{j\in\mathbb{P}_i}}\frac{exp{\left(sim\left(Z_i^{hv},Z_j^{sv}\right)/\tau\right)}}{\mathrm{\Sigma}_{k\in\{\mathbb{P}_i\bigcup{\mathbb{N}_i\}}}exp{\left(sim\left(Z_i^{hv},Z_k^{sv}\right)/\tau\right)}}
\end{equation}

Finally, the overall objective function is:
\begin{equation}
	\mathcal{L}=\frac{1}{\left|V\right|}\sum_{i\in V}\left[\lambda\cdot\mathcal{L}_i^{sv}+\left(1-\lambda\right)\cdot\mathcal{L}_i^{hv}\right]
\end{equation}
where $\lambda$ is the balance coefficient between the graph schema view and hops view. By continuously optimizing the above objective function through the backpropagation algorithm, we can accomplish cross-view self-supervised heterogeneous graph representation learning. During the inference stage for downstream tasks, we select $Z_i^{hv}$ as the final node representation. Because $Z_i^{hv}$ can not only absorb high-quality local information from GNNs under graph schema view through cross-view contrastive learning but also capture long-range global information through Transformer's strong multi-hop feature aggregation ability. Meanwhile, as mentioned above, the hops views branch can conduct inference in a mini-batch manner more conveniently.

\begin{algorithm*}
\caption{GTC: GNN-Transformer Co-contrastive learning algorithm.}
\label{alg:1}
\SetKwInOut{Input}{Input}
\SetKwInOut{Output}{Output}

\Input
{
heterogeneous graph $G=(V,E)$; \\
node feature $x=\left\{x_i,\forall\ i\in V\right\}$;\\
adjacency matrix $A=\left\{A^\varphi,\forall\varphi\in\Phi\right\}$;\\
threshold $\theta_{pos}$ for filtering positive samples;\\
balance coefficient $\lambda$;\\
temperature parameter $\tau$
}
\Output{
The embedding $Z^{hv}$
}

Get projected feature $h=\left\{h_i,\forall\ i\in V\right\}$ by equation(4)\;
\For{$e=1$ \KwTo $epoch$}
{
	Obtain the representation $Z_i^{sv}$ of the node under the Graph Schema view through equation (5) and equation (6) \;
	Obtain the token representation ${S_{\varphi,i}=\left(x_i^0,x_{\varphi,i}^1,\ldots,x_{\varphi,i}^K\right),\forall\ \varphi\in\Phi}$ of different hop neighborhood under different metapaths\;
	Obtain the encoded token representation $Z_{\varphi,v}$ by transformer encoder\;
	Calculate the correlation $\alpha_\varphi^k$ between the node itself and different hop neighborhoods through the Token-level submodule of Hierarchical Attention\;
	Leverage the Token-level submodule of Hierarchical Attention to conduct information aggregation operation, and get the aggregated representation $\{Z_{\varphi,i}, \forall\ i\in V,\forall \varphi \in \Phi \}$ of node $i$ under different metapaths under Hops View\;
	
	Calculate the importance $\alpha_{\varphi}$ of different metapaths by the Semantic-level submodule of Hierarchical Attention\;
	Leverage the Semantic-level submodule of Hierarchical Attention to conduct information fusion operation, and get the final representation $Z_i^{hv}$ of node $i$ under Hops View\;
	Calculate $L_i^{sv}$ and $L_i^{hv}$ by equation (16) and equation (17), respectively\;
	Calculate the overall $\mathcal{L}$ by equation (18).
}
\Return The embedding $Z^{hv}$ of every node.
\end{algorithm*}

\subsection{Analysis of the Proposed Method}\label{sec4.5}

\textbf{Relations with GNNs.} First, as mentioned earlier, the purpose of GTC is to establish collaboratively cross-view contrastive learning between graph schema view and hops view, and solve GNNs' over-smoothing problem. The model structure and information aggregation scheme of the graph schema view branch is not our focus, it can be replaced by any other methodologies of GNNs. Second, from another point of view, GTC can be regarded as an enhanced version of self-supervised GNN, which has the ability to fully capture local and global information without interference from the over-smoothing problem. The existing GNNs are not equipped with this capability, and it is also the biggest advantage of GTC compared with the existing GNNs.

\textbf{Relations with existing Graph Transformers.} Compared with the existing Graph Transformers, the uniqueness of GTC can be summarized as three points: (1) GTC enables efficient feature and structure encoding of heterogeneous graphs without complex operations, such as Laplacian eigenvectors. Compared with NAGphormer\cite{chen2023nagphormer}, it expands the encoding ability on heterogeneous graphs and the aggregation ability of semantic information at different levels. (2) GTC realizes self-supervised heterogeneous graph representation learning. Compared with most existing Graph Transformers which are located in supervised learning paradigms, GTC further improves subsequent applicability in real scenarios. (3) GTC incorporates the powerful local information aggregation ability of GNNs. Although Transformer has excellent long-range and full interaction modeling ability, it is still weaker than GNNs in local information aggregation. GTC can learn from GNNs by establishing cross-view contrastive learning between the Transformer branch and the GNN branch, which can complement its shortcoming in local information aggregation.

In summary, GTC can make optimization for both GNNs and Graph Transformers. At the same time, with the help of the cross-view contrastive learning between GNN and Graph Transformer, their advantages complement each other, and disadvantages are removed.

\section{Experiments}\label{sec5}
In the experiment section, we conduct a lot of experiments on multiple real heterogeneous graph datasets to answer the following questions:

\begin{itemize}
	\item \textbf{RQ1:} How does the proposed method compare with existing state-of-the-art graph representation learning methods, especially self-supervised learning methods?
	\item \textbf{RQ2:} Is the cross-view contrastive learning scheme proposed in this paper between the graph schema view encoded by the GNN branch and the hops view encoded by the Graph Transformer branch effective? Is it beneficial to the performance?
	\item \textbf{RQ3:} We leverage Transformer's ability to capture long-range information and avoid over-smoothing of GNNs while capturing multi-hop neighbors. Can it achieve actually satisfactory performance in eliminating over-smoothing problem?
	\item \textbf{RQ4:} Which hyperparameters have a greater impact on the model performance? How stable is the model under different hyperparameter settings?
\end{itemize}

\subsection{Experimental Settings}\label{sec5.1}

\textbf{Datasets.} In the experiment of this paper, we use the public real heterogeneous graph datasets commonly used in the field, including ACM\cite{zhao2020network}, DBLP\cite{fu2020magnn}, and Freebase\cite{li2021leveraging}. The statistical information of datasets is shown in Table \ref{tab1}.

\begin{table}[htb]
\setlength{\abovecaptionskip}{-0.2cm}
\caption{The basic information of datasets.}
\label{tab1}
\center
\begin{tabular}{c|c|c|c}
\hline
\multicolumn{1}{c|}{Data} & \multicolumn{1}{c|}{Node}                                                                                   & \multicolumn{1}{c|}{Relation}                                                     & \multicolumn{1}{c}{Metapath}                                          \\ \hline
\multicolumn{1}{c|}{ACM}  & \multicolumn{1}{c|}{\begin{tabular}[c]{@{}c@{}}paper(P):4019\\ author(A):7167\\ subject(S):60\end{tabular}} & \multicolumn{1}{c|}{\begin{tabular}[c]{@{}c@{}}P-A:13407\\ P-S:4019\end{tabular}} & \multicolumn{1}{c}{\begin{tabular}[c]{@{}c@{}}PAP\\ PSP\end{tabular}} \\ \hline
DBLP                       & \begin{tabular}[c]{@{}c@{}}author(A):4057\\ paper(P):14328\\ conference(C):20\\ term(T):7723\end{tabular}   & \begin{tabular}[c]{@{}l@{}}P-A:19645\\ P-C:14328\\ P-T:85810\end{tabular}         & \begin{tabular}[c]{@{}c@{}}APA\\ APCPA\\ APTPA\end{tabular}            \\ \hline
Freebase                   & \begin{tabular}[c]{@{}c@{}}movie(M):3492\\ actor(A):33401\\ director(D):2502\\ writer(W):4459\end{tabular}  & \begin{tabular}[c]{@{}c@{}}M-A:65341\\ M-D:3762\\ M-W:6414\end{tabular}           & \begin{tabular}[c]{@{}c@{}}MAM\\ MDM\\ MWM\end{tabular}                \\ \hline
\end{tabular}
\end{table}

\begin{table*}[t]
\setlength{\abovecaptionskip}{-0.3cm} 
  \caption{Comparison results of node classification experiment(\%). Bold for "the best", and underline for "the second best".}
  \label{tab2}
  \setlength\tabcolsep{4pt}
  \center
\begin{tabular}{cccccccccccc}
 \toprule
Datasets                  & Metric                 & Train(Linear) & GraphSAGE & Mp2vec    & HetGNN    & DGI       & HAN       & DMGI      & HeCo      & HeCo++    & \textbf{GTC}       \\
\midrule
\multirow{9}{*}{ACM}      & \multirow{3}{*}{Ma-F1} & 20            & 46.83±4.5 & 52.32±0.8 & 71.89±0.7 & 79.45±3.6 & 85.26±2.3 & 87.86±0.2 & 88.42±0.9 & \underline{89.12±0.6} & \textbf{90.20±0.7} \\
                          &                        & 40            & 56.21±6.2 & 63.27±0.9 & 71.61±0.5 & 80.23±3.3 & 87.47±1.1 & 86.23±0.8 & 87.61±0.5 & \underline{88.70±0.7} & \textbf{88.92±0.6} \\
                          &                        & 60            & 56.59±5.7 & 61.13±0.4 & 74.33±0.6 & 80.03±3.3 & 88.41±1.1 & 87.97±0.4 & 89.04±0.5 & \underline{89.51±0.7} & \textbf{89.91±0.2} \\
                          \cmidrule(l){2-12}
                          & \multirow{3}{*}{Mi-F1} & 20            & 49.72±5.5 & 53.13±0.9 & 71.89±1.1 & 79.63±3.5 & 85.11±2.2 & 87.60±0.8 & 88.13±0.8 & \underline{88.96±0.5} & \textbf{90.64±0.2} \\
                          &                        & 40            & 60.98±3.5 & 64.43±0.6 & 74.46±0.8 & 80.41±3.0 & 87.21±1.2 & 86.02±0.9 & 87.45±0.5 & \underline{88.40±0.8} & \textbf{88.55±0.3} \\
                          &                        & 60            & 60.72±4.3 & 62.72±0.3 & 76.08±0.7 & 80.15±3.2 & 88.10±1.2 & 87.82±0.5 & 88.71±0.5 & \underline{89.30±0.7} & \textbf{89.45±0.4} \\
                           \cmidrule(l){2-12}
                          & \multirow{3}{*}{AUC}   & 20            & 65.88±3.7 & 71.66±0.7 & 84.36±1.0 & 91.47±2.3 & 93.47±1.5 & 96.72±0.3 & 96.49±0.3 & \underline{97.25±0.2} & \textbf{97.58±0.1} \\
                          &                        & 40            & 71.06±5.2 & 80.48±0.4 & 85.01±0.6 & 91.52±2.3 & 94.84±0.9 & 96.35±0.3 & 96.40±0.4 & \underline{97.08±0.2} & \textbf{97.54±0.2} \\
                          &                        & 60            & 70.45±6.2 & 79.33±0.4 & 87.64±0.7 & 91.41±1.9 & 94.68±1.4 & 96.79±0.2 & 96.55±0.3 & \underline{97.50±0.2} & \textbf{97.82±0.1} \\
                           \midrule 
\multirow{9}{*}{DBLP}     & \multirow{3}{*}{Ma-F1} & 20            & 71.97±8.4 & 88.98±0.2 & 89.51±1.1 & 87.93±2.4 & 89.31±0.9 & 89.94±0.4 & 91.28±0.2 & \underline{91.40±0.2} & \textbf{93.12±0.3} \\
                          &                        & 40            & 73.69±8.4 & 88.68±0.2 & 88.61±0.8 & 88.62±0.6 & 88.87±1.0 & 89.25±0.4 & 90.34±0.3 & \underline{90.56±0.2} & \textbf{91.62±0.3} \\
                          
                          &                        & 60            & 73.86±8.1 & 90.25±0.1 & 89.56±0.5 & 89.19±0.9 & 89.20±0.8 & 89.46±0.6 & 90.64±0.3 & \underline{91.01±0.3} & \textbf{92.95±0.2} \\
                          \cmidrule(l){2-12}
                          & \multirow{3}{*}{Mi-F1} & 20            & 71.44±8.7 & 89.67±0.1 & 90.11±1.0 & 88.72±2.6 & 90.16±0.9 & 90.78±0.3 & 91.97±0.2 & \underline{92.03±0.1} & \textbf{93.67±0.3} \\
                          &                        & 40            & 73.61±8.6 & 89.14±0.2 & 89.03±0.7 & 89.22±0.5 & 89.47±0.9 & 89.92±0.4 & 90.76±0.3 & \underline{90.87±0.2} & \textbf{92.02±0.3} \\
                          &                        & 60            & 74.05±8.3 & 91.17±0.1 & 90.43±0.6 & 90.35±0.8 & 90.34±0.8 & 90.66±0.5 & 91.59±0.2 & \underline{91.86±0.2} & \textbf{93.61±0.2} \\
                          \cmidrule(l){2-12}
                          & \multirow{3}{*}{AUC}   & 20            & 90.59±4.3 & 97.69±0.0 & 97.96±0.4 & 96.99±1.4 & 98.07±0.6 & 97.75±0.3 & 98.32±0.1 & \underline{98.39±0.1} & \textbf{98.96±0.1} \\
                          &                        & 40            & 91.42±4.0 & 97.08±0.0 & 97.70±0.3 & 97.12±0.4 & 97.48±0.6 & 97.23±0.2 & 98.06±0.1 & \underline{98.17±0.1} & \textbf{98.46±0.1} \\
                          &                        & 60            & 91.73±3.8 & 98.00±0.0 & 97.97±0.2 & 97.76±0.5 & 97.96±0.5 & 97.72±0.4 & 98.59±0.1 & \underline{98.62±0.1} & \textbf{98.89±0.1} \\
                          \midrule 
\multirow{9}{*}{Freebase} & \multirow{3}{*}{Ma-F1} & 20            & 45.14±4.5 & 53.96±0.7 & 52.72±1.0 & 54.90±0.7 & 53.16±2.8 & 55.79±0.9 & 59.23±0.7 & \underline{59.87±1.0} & \textbf{60.40±1.5} \\
                          &                        & 40            & 44.88±4.1 & 57.80±1.1 & 48.57±0.5 & 53.40±1.4 & 59.63±2.3 & 49.88±1.9 & \underline{61.19±0.6} & \textbf{61.33±0.5} & 60.20±0.9 \\
                          &                        & 60            & 45.16±3.1 & 55.94±0.7 & 52.37±0.8 & 53.81±1.1 & 56.77±1.7 & 52.10±0.7 & 60.13±1.3 & \textbf{60.86±1.0} & \underline{60.81±1.2} \\
                          \cmidrule(l){2-12}
                          & \multirow{3}{*}{Mi-F1} & 20            & 54.83±3.0 & 56.23±0.8 & 56.85±0.9 & 58.16±0.9 & 57.24±3.2 & 58.26±0.9 & 61.72±0.6 & \underline{62.29±1.9} & \textbf{64.58±1.7} \\
                          &                        & 40            & 57.08±3.2 & 61.01±1.3 & 53.96±1.1 & 57.82±0.8 & 63.74±2.7 & 54.28±1.6 & 64.03±0.7 & \underline{64.27±0.5} & \textbf{64.90±1.6} \\
                          &                        & 60            & 55.92±3.2 & 58.74±0.8 & 56.84±0.7 & 57.96±0.7 & 61.06±2.0 & 56.69±1.2 & 63.61±1.6 & \underline{64.15±0.9} & \textbf{65.86±1.3} \\
                          \cmidrule(l){2-12}
                          & \multirow{3}{*}{AUC}   & 20            & 67.63±5.0 & 71.78±0.7 & 70.84±0.7 & 72.80±0.6 & 73.26±2.1 & 73.19±1.2 & \underline{76.22±0.8} & \textbf{76.68±0.7} & 75.21±0.9 \\
                          &                        & 40            & 66.42±4.7 & 75.51±0.8 & 69.48±0.2 & 72.97±1.1 & 77.74±1.2 & 70.77±1.6 & \underline{78.44±0.5} & \textbf{79.51±0.3} & 77.10±1.5 \\
                          &                        & 60            & 66.78±3.5 & 74.78±0.4 & 71.01±0.5 & 73.32±0.9 & 75.69±1.5 & 73.17±1.4 & \underline{78.04±0.4} & \textbf{78.27±0.7} & 76.25±1.4\\
             
                          \midrule

\multirow{9}{*}{Academic} & \multirow{3}{*}{Ma-F1} & 20            & 45.14±4.5 & 53.96±0.7 & 52.72±1.0 & 54.90±0.7 & 78.08±0.2 & 55.79±0.9 & 59.23±0.7 & \underline{59.87±1.0} & \textbf{60.40±1.5} \\
                          &                        & 40            & 44.88±4.1 & 57.80±1.1 & 48.57±0.5 & 53.40±1.4 & 59.63±2.3 & 49.88±1.9 & \underline{61.19±0.6} & \textbf{61.33±0.5} & 60.20±0.9 \\
                          &                        & 60            & 45.16±3.1 & 55.94±0.7 & 52.37±0.8 & 53.81±1.1 & 56.77±1.7 & 52.10±0.7 & 60.13±1.3 & \textbf{60.86±1.0} & \underline{60.81±1.2} \\
                          \cmidrule(l){2-12}
                          & \multirow{3}{*}{Mi-F1} & 20            & 54.83±3.0 & 56.23±0.8 & 56.85±0.9 & 58.16±0.9 & 75.06±0.2 & 58.26±0.9 & 61.72±0.6 & \underline{62.29±1.9} & \textbf{64.58±1.7} \\
                          &                        & 40            & 57.08±3.2 & 61.01±1.3 & 53.96±1.1 & 57.82±0.8 & 63.74±2.7 & 54.28±1.6 & 64.03±0.7 & \underline{64.27±0.5} & \textbf{64.90±1.6} \\
                          &                        & 60            & 55.92±3.2 & 58.74±0.8 & 56.84±0.7 & 57.96±0.7 & 61.06±2.0 & 56.69±1.2 & 63.61±1.6 & \underline{64.15±0.9} & \textbf{65.86±1.3} \\
                          \cmidrule(l){2-12}
                          & \multirow{3}{*}{AUC}   & 20            & 67.63±5.0 & 71.78±0.7 & 70.84±0.7 & 72.80±0.6 & 95.42±0.1 & 73.19±1.2 & \underline{76.22±0.8} & \textbf{76.68±0.7} & 75.21±0.9 \\
                          &                        & 40            & 66.42±4.7 & 75.51±0.8 & 69.48±0.2 & 72.97±1.1 & 77.74±1.2 & 70.77±1.6 & \underline{78.44±0.5} & \textbf{79.51±0.3} & 77.10±1.5 \\
                          &                        & 60            & 66.78±3.5 & 74.78±0.4 & 71.01±0.5 & 73.32±0.9 & 75.69±1.5 & 73.17±1.4 & \underline{78.04±0.4} & \textbf{78.27±0.7} & 76.25±1.4\\
             
                          \midrule 
  \vspace{-0.6cm}

\end{tabular}

\end{table*}

\textbf{Baselines.} To fully verify the performance of the method proposed in this paper, in the experiment, we compared GTC with the existing state-of-the-art methods. These methods can be divided into: (1) unsupervised homogeneous methods, such as GraphSAGE\cite{hamilton2017inductive} and DGI\cite{velickovic2019deep}; (2) unsupervised heterogeneous methods, such as Mp2vec\cite{dong2017metapath2vec}, HetGNN\cite{zhang2019heterogeneous}, DMGI\cite{park2020unsupervised}, HeCo\cite{wang2021self} and HeCo++\cite{liu2023hierarchical}; (3) the semi-supervised heterogeneous method HAN\cite{wang2019heterogeneous}. All the results in the experiment are achieved using the official open-source code.

\textbf{Implementation details.} First, for homogeneous methods such as GraphSAGE, and DGI, according to the common methods in the field, we utilize the metapaths shown in Table \ref{tab1} to extract different homogeneous graphs to conduct performance tests respectively, and finally select the best one as the final result of these methods. For some methods based on random walks, referring to HeCo++\cite{liu2023hierarchical}, we set the number of random walks for each node to 40, the path length of the walks to 100, and the window size to 5. For HAN, both the number and dimension of heads are set to 8. For other parameter settings, we follow the original papers.

For our proposed GTC, we leverage Adam to optimize the model parameters. When determining the hyperparameter settings of the model, we adopt the grid search strategy. Specifically, for the layer number $L_g$ of the GNN branch, we search it within the range $\left[1,6\right]$, for the layer number $L_{tm}$ of the Transformer branch, headers of MSA, and max hop $K$ of Metapath-aware Hop2Token, we search them within the range $\left[1,9\right]$. The learning rate is searched according to the uniform distribution strategy within the range $\left[0.0001,0.01\right]$, The temperature parameter $\tau$ and balance coefficient $\lambda$ are searched from 0.1 to 0.9 with step 0.1. In addition, we adopt the early stop strategy, and the patience is set to 30. 
 For all methods, we uniformly set the final node embedding dimension to 64. To ensure the stability, comprehensiveness, and credibility of the results, all the experimental results in this paper are the average of 10 experimental results. Our code is built upon HeCo\cite{wang2021self}, NAGphormer\cite{chen2023nagphormer}, and OpenHGNN\cite{han2022openhgnn}.



\subsection{Performance Comparison (RQ1)}\label{sec5.2}

\subsubsection{Node Classification}\label{sec5.2.1}

First, we compare the performance of different methods on the node classification task. Specifically, we first obtain the embedding of nodes using different baseline methods described in \ref{sec5.1} and our proposed method. Then, a linear classifier is trained based on these embeddings. Following the experimental settings of HeCo++\cite{liu2023hierarchical}, we randomly select 20, 40, and 60 labeled nodes in each category as the training set of the linear classifier. We choose 1000 nodes in each category as the verification and the test set, respectively. 
 To more comprehensively verify the node classification performance of different methods, we selected the commonly used evaluation metrics in the field, including Ma-F1, Mi-F1, and AUC.
\begin{figure*}[t]
  \centering
  \includegraphics[width=7in,trim=0 20 0 0]{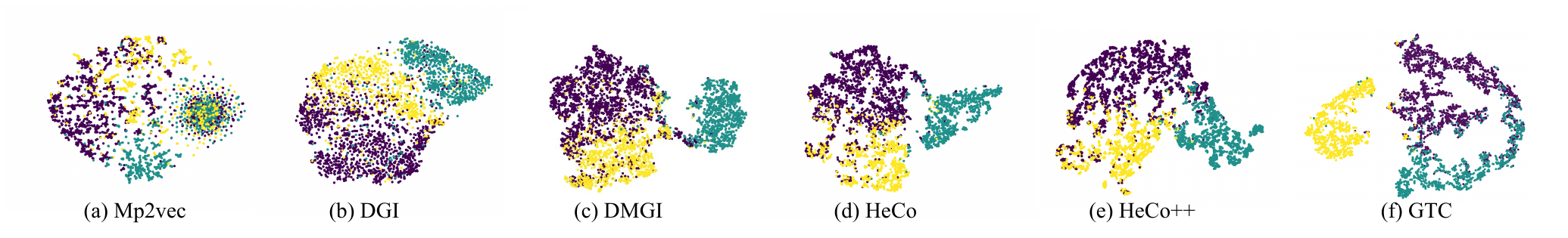}
  \caption{Visualization of the learned node embedding on ACM. The Silhouette scores for (a), (b), (c), (d), (e), (f) are 0.0292, 0.1862, 0.3015, 0.3642, 0.3885 and 0.4327, respectively.}
  \label{fig3}
  \vspace{-0.5cm}
  \end{figure*}
The node classification performance obtained by each method is shown in Table \ref{tab2}. It can be seen from the table that the GTC proposed by us achieves the best or most competitive performance under all training data splits on all datasets. Because it fully utilizes the cross-view collaboratively contrastive learning of GNN and Graph Transformer to realize the comprehensive capture of local and global information. It is particularly worth noting that although the semi-supervised method HAN has known the node labels in the training set in advance and has been guided by validation in the process of learning node representation, it does not achieve the best performance as expected. The reason is that the semi-supervised training paradigm makes it heavily relies on the supervision signal of node labels. When the training data is relatively sufficient, such as more than 20\%  labeled nodes of the entire dataset, it may achieve satisfactory performance. However, in this paper, when we limit the number of training sets to 20, 40, or 60 nodes per category, the amount of supervision information that the model can leverage is extremely limited, so its performance is not satisfactory. In real scenarios, labeling nodes is an extremely time-consuming and laborious task. Therefore, the experimental settings in this paper can better reflect the real application scenarios, which also poses a huge challenge to the practical application of some supervised methods.
  
We then compare the homogeneous approaches with the heterogeneous ones. Although the performance of the homogeneous methods in Table \ref{tab2} is the best result selected from its performance on all metapaths, we can see that the performance of the homogeneous methods is generally inferior to that of the heterogeneous methods. This indicates that a single metapath cannot fully reflect all the semantic information of a heterogeneous graph, even if the metapath is carefully selected. Among all homogeneous methods, DGI achieves significantly better results than other methods, and surprisingly, even outperforms some heterogeneous methods (Mp2vec, HetGNN). The reason is that DGI maximizes the mutual information of the high-order global and the local representation, enabling it to capture the global information of the graph, which also illustrates the importance of the global information of the graph. DMGI makes an extension of supporting different relationships in heterogeneous graphs, which makes its performance significantly improved compared to DGI, which can also reflect the necessity and value of mining different semantic relationships in heterogeneous graphs. HeCo and HeCo++ outperform DMGI in all cases, which shows the superiority of cross-view to single-view. By establishing contrastive learning across views, richer and more comprehensive information can be captured, which can also be well verified in the variant experiments in section \ref{sec5.3} of this paper. The GTC proposed in this paper can not only capture the multi-hop global semantic information under the heterogeneous relationship through Metapath-aware Hop2Token, but also realize the attentive fusion of different level semantic information through Hierarchical Attention. At the same time, GTC can also better mine multi-view information through the cross-view contrastive learning of graph schema view and hops view. These excellent characteristics make it possible for GTC to achieve the best performance.

\subsubsection{Node Clustering}\label{sec5.2.2}

In this section, we feed the node embedding obtained by different methods into the K-means model to conduct the node clustering task. We select widely adopted metrics to evaluate the quality of node clustering, including normalized mutual information (NMI), and adjusted rand index (ARI). Since the semi-supervised method HAN has known the node labels in training set in advance and has been guided by validation in the process of learning node representation, we remove it in this task. The experimental results are shown in Table \ref{tab3}. 

From the table, we can see that the performance is roughly similar to that in the node classification task. GTC consistently outperforms other baselines by a large margin, except for being slightly inferior to HeCo++ on Freebase. This implies that the node embedding learned by GTC has great generality for various downstream tasks. In addition, some similar and meaningful phenomena as in the node classification task still appear: The fusion of semantic information from different metapaths promotes heterogeneous methods to generally outperform homogeneous methods; the capture of global information makes DGI exhibit superiority to heterogeneous methods such as Mp2vec and HetGNN; cross-view contrastive learning brings benefits to HeCo and HeCo++, which can outperform single-view DMGI in all cases. This once again proves the great potential of the GTC with all the above excellent qualities.

\begin{table*}[t]
\setlength{\abovecaptionskip}{-0.2cm}
\caption{Comparison results of node clustering experiment(\%).Bold for "the best", and underline for "the second best".}
\label{tab3}
\center
\begin{tabular}{c|cc|cc|cc|cc}
\hline
Datasets                          & \multicolumn{2}{c|}{ACM}        & \multicolumn{2}{c|}{DBLP}       & \multicolumn{2}{c}{Freebase} & \multicolumn{2}{c}{Academic} \\ \hline
Metrics                           & NMI            & ARI            & NMI            & ARI            & NMI           & ARI      & NMI           & ARI     \\ \hline
GraphSage & 29.2           & 27.72          & 51.5           & 36.4           & 9.05          & 10.49        \\
Mp2vec                            & 48.43          & 34.65          & 73.55          & 77.7           & 16.47         & 17.32        \\
HetGNN                            & 41.53          & 34.81          & 69.79          & 75.34          & 12.25         & 15.01        \\
DGI       & 51.73          & 41.16          & 59.23          & 61.85          & 18.34         & 11.29        \\
DMGI                              & 51.66          & 46.64          & 70.06          & 75.46          & 16.98         & 16.91   
\\
HeCo                              & 56.87          & 56.94          & 74.51          & 80.17          & \underline{20.38}         & 20.98     \\

HeCo++                            & \underline{60.82}          & \underline{60.09}          & \underline{75.39}          & \underline{81.2}           & \textbf{20.62}         & \textbf{21.88}        \\ 
\textbf{GTC}                               & \textbf{63.26} & \textbf{69.30} & \textbf{78.10} & \textbf{82.89} & 18.70   & \underline{21.31}  \\
\hline
\end{tabular}
\end{table*}

\subsubsection{Visualization Experiment}\label{sec5.2.3}

Considering that the distribution quality of node embedding directly determines the performance of downstream tasks, in this section, to more intuitively display the distribution of node embedding learned by different models, we conduct a visualization experiment. Following the most widely-used scheme in this field, we first use the t-SNE\cite{van2008visualizing} to compress the learned node embeddings into 2-dimensional space, and then visualize them. We select some representative results on the ACM dataset to display in Fig.\ref{fig3}, where color represents the nodes categories. In addition, to numerically compare the distribution quality, we introduce Silhouette scores to quantify the outline clarity of each category after clustering. 

From Fig.\ref{fig3}, we can intuitively see that except for Mp2vec in Fig.\ref{fig3}(a), the spatial distribution of node embedding in Fig.\ref{fig3}(b)-(f) has a strong correlation with the labels, which means that the methods in Fig.\ref{fig3}(b)-(f) have learned node embeddings with the relatively fine spatial distribution. In Fig.\ref{fig3}(a), there are intersecting phenomena between different category nodes, and the Silhouette score is far lower than other methods, which can also well explain why Mp2vec lags behind other methods in both node classification and node clustering.

Next, let's focus on the boundaries of different node clusters. It is widely recognized that distinct boundary separation can reduce the learning difficulty of downstream task models so that very simple models can also achieve excellent downstream task performance. By observing Fig.\ref{fig3}(b)-(f), we can easily find their differences. Although DGI can gather nodes of the same type into a cluster, there present blurred boundaries between different clusters. The situation of DMGI, HeCo, and HeCo++ is relatively better. The green nodes in the three subgraphs can be well separated from the other two clusters, but there are still some intersections between the purple and yellow nodes. In contrast, the distribution of node embeddings obtained by GTC is optimal among all methods. Not only is the spatial distribution of node embeddings highly correlated with labels but also it can achieve higher inter-cluster dispersion and intra-cluster concentration. This implies that GTC can not only provide great support for superior downstream task performance but also greatly reduce the downstream processing difficulty and model complexity. Based on the above analysis, we can draw a conclusion that GTC has the potential to achieve considerable performance on more downstream tasks.

\subsection{Variant Analysis (RQ2)}\label{sec5.3}
\begin{figure*}[t]
  \centering
  \includegraphics[width=6.5in,trim=0 20 0 20]{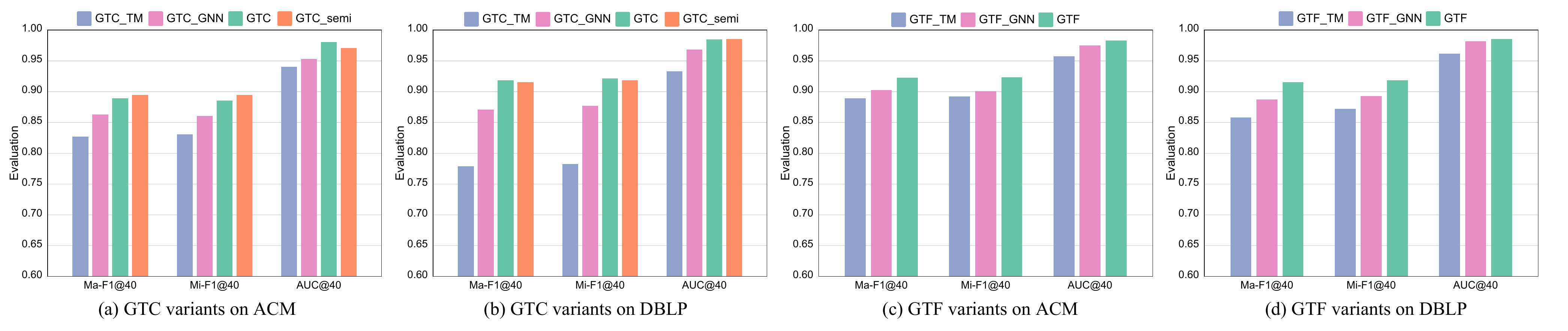}
  \caption{The node classification performance comparison of different variants of GTC and GTF on ACM and DBLP datasets.}
  \label{fig4}
  \vspace{-0.5cm}
  \end{figure*}
In this section, we verify the effectiveness of cross-view contrastive learning between the graph schema view and hops view, and the feasibility of the information fusion scheme of GNN and Graph Transformer, by constructing different model variants. Without loss of generality, in this experiment, we select 40 labeled nodes in each category as the training data of the linear classifier. For other information, please refer to section \ref{sec5.2.1}.

Specifically, we first construct two variants on the basis of GTC: GTC\_TM and GTC\_GNN, which perform contrastive learning only within the hops view or graph schema view, respectively. We expect to show the benefits of cross-view contrastive learning by comparing the performance of GTC\_TM, GTC\_GNN, and GTC. At the same time, we use the labeled training set nodes (the same data as the subsequent training of the linear classifier) to fine-tune GTC on the node classification task and obtain the corresponding variant GTC\_semi. By comparing GTC\_semi and GTC, we can verify whether limited supervision information can boost the performance of GTC. 

The performance comparison results are shown in Fig.\ref{fig4}(a) and Fig.\ref{fig4}(b). It can be seen that GTC  consistently outperforms the two single-view variants GTC\_TM and GTC\_GNN by a large margin, which is enough to illustrate the effectiveness of the proposed cross-view contrastive learning scheme between graph schema view and hops view, which is also consistent with the analysis in section \ref{sec5.2.1} and \ref{sec5.2.2}. The performance of GTC\_TM is obviously inferior to that of GTC\_GNN, the reason is that Transformer is weaker than GNN in local information aggregation ability. Local information is very important for graph node representation and the tokens alone with MSA cannot completely replace the message-passing of GNNs. This is in line with the previous analysis and expectations of this paper. It is also the reason why this paper chooses Graph Transformer and GNN for collaboratively contrastive learning to complement each other, instead of completely replacing GNN with Graph Transformer. At the same time, we also find that GTC\_semi does not show prefer performance to GTC.
This demonstrates that limited supervision information cannot significantly improve the performance of GTC, which is also consistent with the analysis of HAN in section \ref{sec5.2.1} of this paper.

In addition, to better explore the complementary effects of the collaborative fusion of the global information encoded by Graph Transformer and the local information encoded by GNN, we modify the GTC as shown in Fig.\ref{fig1}(a) and construct the GNN-Transformer Fusion learning model (GTF). The specific modifications mainly include: (1) Removing the 1 to $L_g$ hops of GTC that have been encoded in the graph schema view from the hops view branch, that is, Transformer only encodes the global information; (2)The contrastive learning task is replaced by the node classification task, which takes the concatenation of the local view and global view embedding as the input, so GTF evolves as a semi-supervised method. Similarly, we also construct two variants on the basis of GTF: GTF\_TM and GTF\_GNN, which only rely on the global view or local view, respectively. The performance results are shown in Fig.\ref{fig4}(c) and Fig.\ref{fig4}(d). It can be seen that GTF shows superiority to its variants in all cases, which indicates that the fusion of local view and global view information brings performance improvement. This is in line with the original insights of this paper. Similar to the phenomenon of Fig.\ref{fig4}(a) and Fig.\ref{fig4}(b), the performance of GTF\_TM is inferior to that of GTF\_GNN, which is consistent with the previous analysis.

\subsection{Exploration of GTC's Resistance Ability to Over-smoothing (RQ3)}\label{sec5.4}

The uniqueness of this paper is to construct cross-view contrastive learning between the graph schema view and hops view to eliminate over-smoothing while capturing multi-hop neighbors. To verify the ability of GTC to resist the over-smoothing problem under the condition of capturing as much hop neighbor information as possible, we quantitatively and qualitatively analyze the performance of GTC, GTF, and RGCN under different \#layers/hops settings. Specifically, for RGCN, \#layers/hops refers to the layers of networks, for GTC and GTF, \#layers/hops refers to the maximum hop $K$.

\begin{figure}[htb]
  \centering
  \includegraphics[width=3.5in,trim=0 0 0 0]{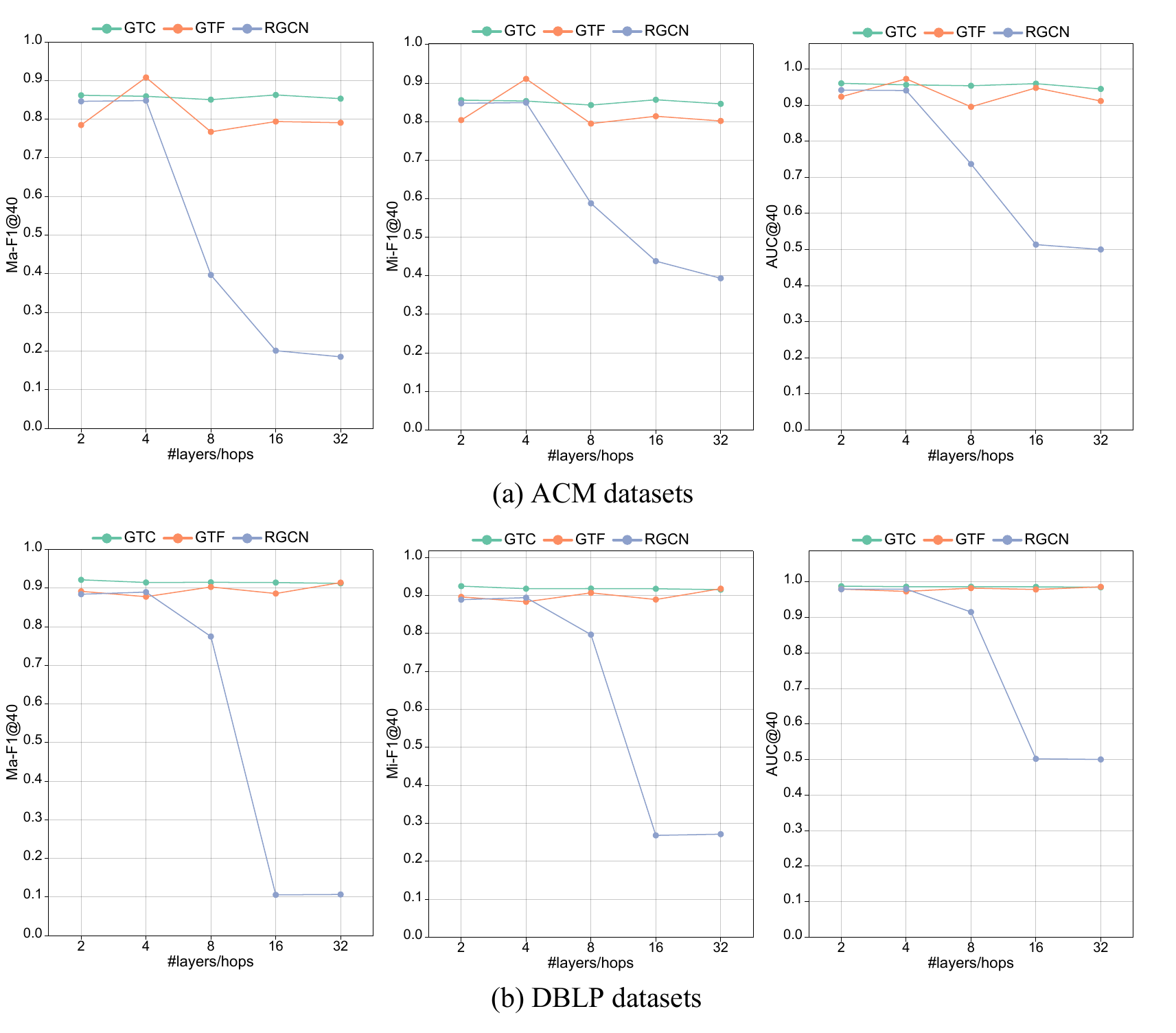}
  \caption{Quantitative performance analysis of different methods with different \#layers/hops settings on different datasets.}
  \label{fig5}
  \end{figure}

Fig.\ref{fig5}(a) and Fig.\ref{fig5}(b) show the quantitative performance of the three methods on the ACM and DBLP datasets, respectively. It intuitively shows that when \#layers/hops is increased to more than 4, the performance of RGCN on both datasets begins to decline dramatically. This indicates that RGCN suffers from over-smoothing problems severely, which is consistent with the phenomenon in\cite{li2019deepgcns,li2020deepergcn}. Surprisingly, GTC and GTF proposed in this paper can still maintain high performance, even if \#layers/hops reach 32, which shows that the two methods possess the excellent capacity to resist the over-smoothing problem. Although increasing \#layers/hops do not significantly promote the performance of GTC and GTF on the datasets in this paper, we believe that for some tasks or datasets that need to capture multi-hop neighbors, our model's advantage will be better demonstrated.

  \begin{figure*}[t]
  \centering
  \includegraphics[width=6in,trim=0 20 0 0]{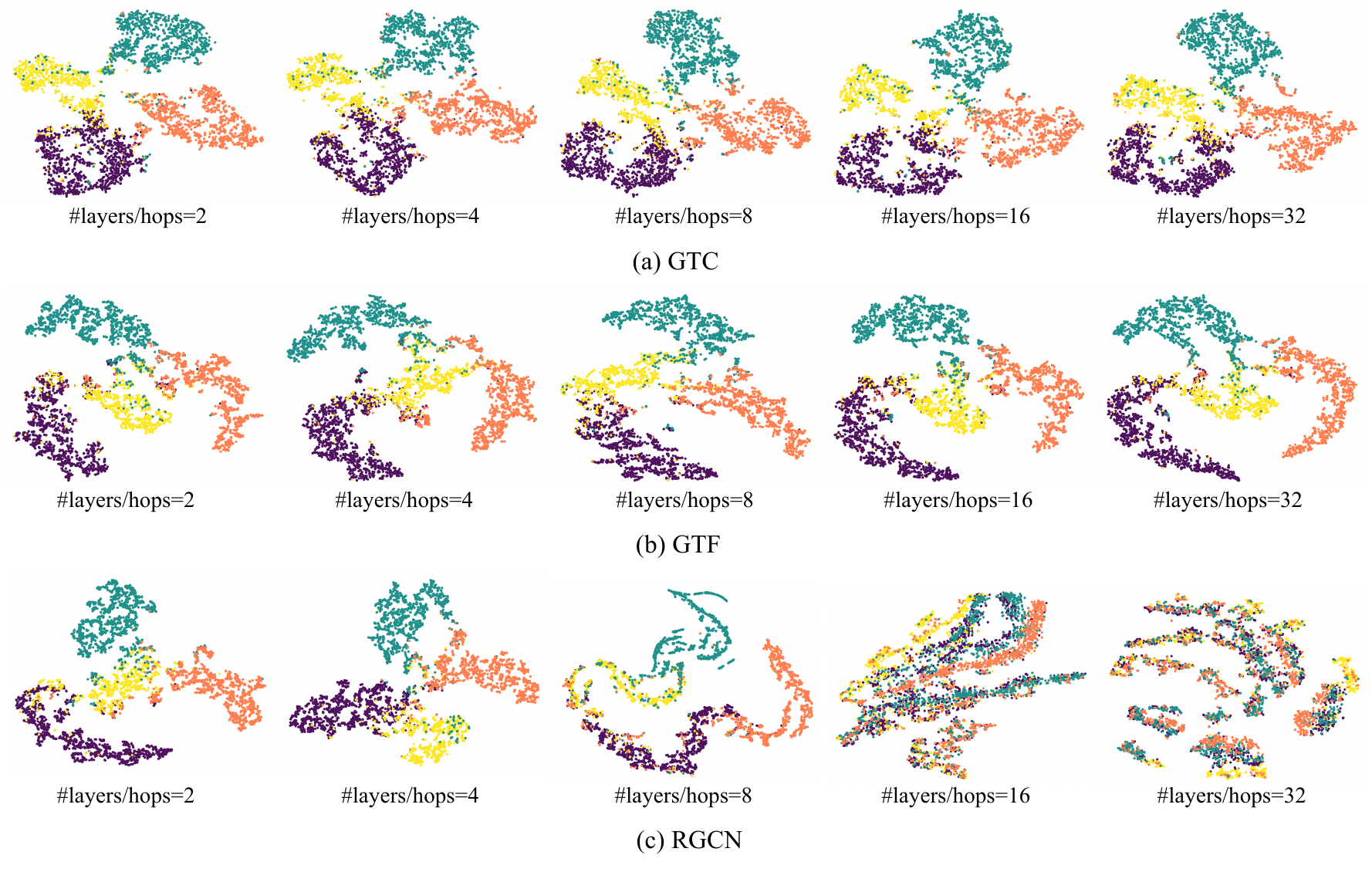}
  \caption{t-SNE visualization of node representations derived from different methods with different \#layers/hops settings on the DBLP dataset.}
  \label{fig6}
\vspace{-0.4cm}
  \end{figure*}

Correspondingly, we also conduct a qualitative analysis of the representation performance. Fig.\ref{fig6} shows the visual distribution of node embeddings obtained by the three methods under different \#layers/hops settings. In Fig.\ref{fig6}(a) and Fig.\ref{fig6}(b), with the increase of \#layers/hops, the distribution of nodes does not fluctuate, maintaining a very considerable distinguishability. However, in Fig.\ref{fig6}(c), when \#layers/hops exceed 4, the node distribution becomes confusing and nodes cannot be distinguished, which is caused by RGCN's over-smoothing. The above experimental phenomena are sufficient to illustrate the excellent performance of our methods in resisting over-smoothing and can also well explain the phenomena in Fig.\ref{fig5}.

\subsection{Hyper-parameter Studies of GTC (RQ4)}\label{sec5.5}

\begin{figure}[htb]
\setlength{\abovecaptionskip}{-0.2cm}
  \centering
  \includegraphics[width=3.5in,trim=0 10 0 20]{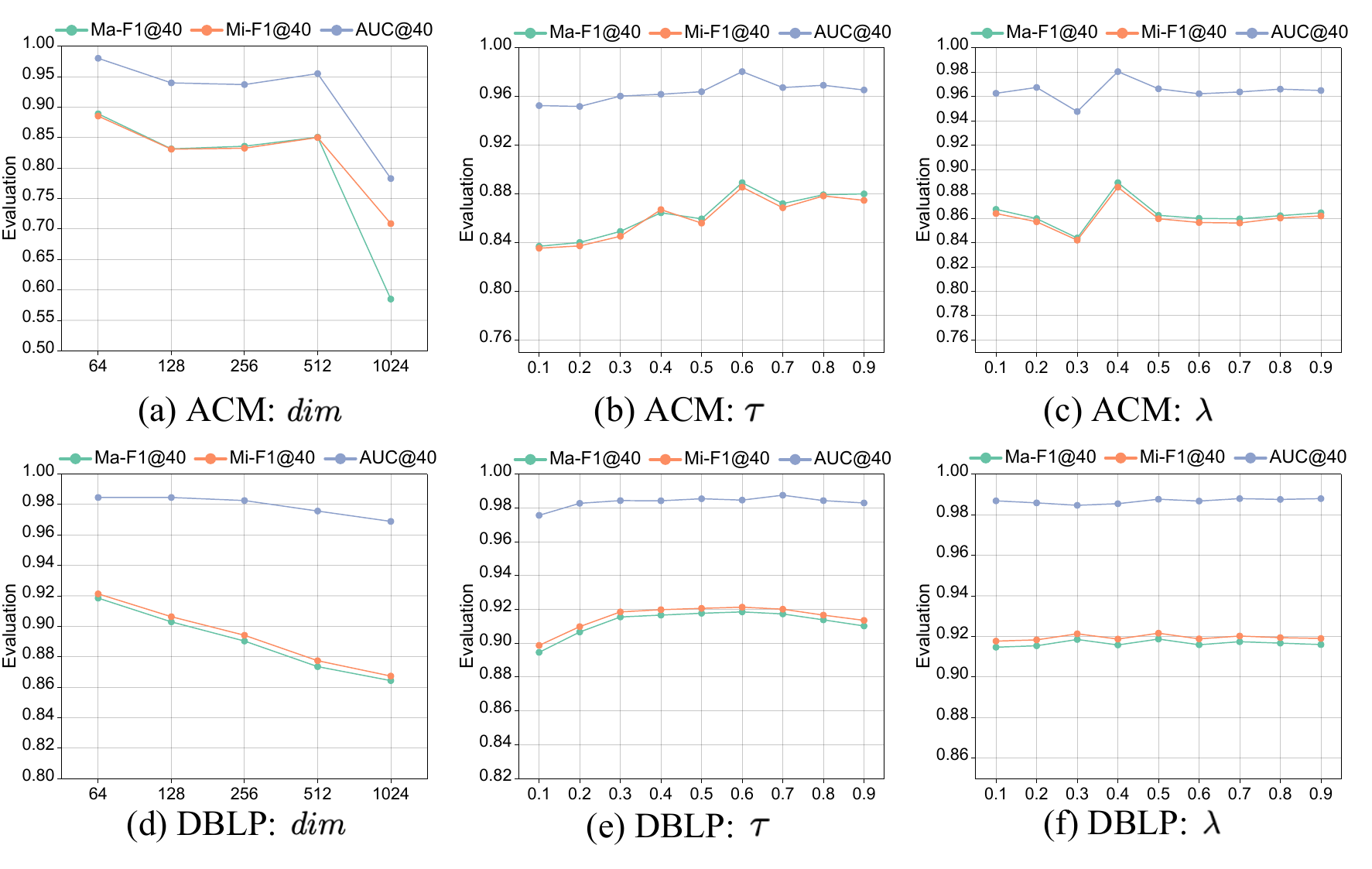}
  \caption{The performance of GTC with different $dim$, $\tau$ and $\lambda$ settings on different datasets.}
  \label{fig8}
  \end{figure}
%

%
Generally speaking, the higher dimension of node embedding, the richer information it contains, and the better downstream task performance. However, by observing Fig.\ref{fig8}(a) and Fig.\ref{fig8}(d), we find a surprising phenomenon: As $dim$ increases, the performance of GTC shows a declining trend on both datasets. We analyze that the reason is that when $dim=64$, the model has been able to mine enough rich information, and we increase $dim$ continually, the node embedding is mixed with irrelevant information, which leads to a decline in model performance. On the other hand, considering the complexity of the model and downstream tasks, and the storage space of node embedding, setting $dim=64$ is also the best choice. $\tau$ determines the smoothness of similarity measures of different embedding spaces, the smaller $\tau$ means sharper similarity, and the larger $\tau$ means smoother similarity. The correct $\tau$ setting can make the model learn hard negatives better. From Fig.\ref{fig8}(b) and Fig.\ref{fig8}(e), it can be seen that with the increase of $\tau$, the performance of the model first shows a rising trend until reaches saturation, and then it will show a slight downward tendency, when $\tau=0.6$, the model achieves the best performance. $\lambda$ has little influence on the model performance, and $\lambda=0.4$ is a reasonable choice.

\section{Conclusion and Future Works}

Facing the over-smoothing problem that has troubled GNNs for a long time, this paper innovatively proposes a GNN-Transformer collaboratively contrastive learning scheme for the first time. Based on this, we propose the GTC architecture, in which the GNN and Transformer are leveraged as two branches to encode node information from different views respectively, and the contrastive learning task is constructed based on the encoded cross-view information to realize self-supervised heterogeneous graph representation. For the Transformer branch, we first propose Metapath-aware Hop2Token, which realizes the efficient transformation from different hop neighbors to token series in heterogeneous graphs. We also propose the CG-Hetphormer model, which not only attentively fuse Token-level and Semantic-level semantic information but also cooperates with the GNN branch to achieve collaborative learning. Tremendous experiments on the real heterogeneous datasets show that the performance of our proposed method is superior to the existing state-of-the-art methods, which proves that our proposed collaborative learning scheme of GNN and Transformer is effective. In particular, when the model goes deeper, our method can also maintain high performance and stability, this proves that the multi-hop neighbor information can be captured without interference from the over-smoothing problem, which provides a reference for future research on solving the over-smoothing problem of GNNs.


Of course, our method is the first attempt on GNN-Transformer collaboratively contrastive learning, and there is still room for improvement. For example, after Metapath-aware Hop2Token obtains the token sequence, we can try to leverage an elastic computing strategy\cite{yang2020resolution,huang2022glance} to allow the model to dynamically determine the number of hops according to the current computing resources, target tasks, and node features, instead of manually setting. 
 In addition, we can make an attempt to generate harder negatives with the help of GAN\cite{liu2023hierarchical} to further improve the model performance. 

\section*{Acknowledgment}

This work is supported by National Key R\&D Program of China (2020YFE0201500), the Fundamental Research Funds for the Central Universities (HIT.NSRIF.201714), Weihai Science and Technology Development Program (2016DXGJMS15), and the Key Research and Development Program in Shandong Province (2017GGX90103).

 Our code is built upon HeCo\cite{wang2021self}, NAGphormer\cite{chen2023nagphormer}, and OpenHGNN\cite{han2022openhgnn}, we thank the authors for their open-sourced code.

\bibliography{bib}

\end{document}